\documentclass[11pt,a4paper]{article}
\usepackage[hyperref]{emnlp-ijcnlp-2019}
\usepackage{times}
\usepackage{latexsym}
\usepackage{graphicx}
\usepackage{cleveref}
\crefformat{section}{\S#2#1#3}
\usepackage{multirow}
\usepackage{siunitx}
\usepackage{comment}
\usepackage{multicol}
\usepackage{array, caption, tabularx,  ragged2e,  booktabs}
\usepackage{natbib}
\usepackage{amsmath, nccmath}
\usepackage{enumitem}

\usepackage{amssymb}
\usepackage{tabu}
\usepackage{easybmat}
\usepackage{hyperref}
\usepackage{lscape}
\usepackage{comment}
\usepackage[normalem]{ulem}

\usepackage{xcolor}

\usepackage{url}

\aclfinalcopy 

\usepackage{xspace}

\newcommand{\alert}[1]{\textcolor{red}{\noindent$\Rightarrow$ #1}}
\newcommand{\red}[1]{\textcolor{red}{#1}}
\newcommand{\blue}[1]{\textcolor{blue}{#1}}

\newcommand{\eg}{{\em e.g.,}\xspace}

\newcommand{\Ni}{({\em i})~}
\newcommand{\Nii}{({\em ii})~}
\newcommand{\Niii}{({\em iii})~}
\newcommand{\Niv}{({\em iv})~}

\newcommand{\Na}{({\em a})~}
\newcommand{\Nb}{({\em b})~}
\newcommand{\Nc}{({\em c})~}

\newcommand{\wsj}{\textsc{wsj}}
\newcommand{\gridcnn}{\nobreak{Neural Grid}}
\newcommand{\gridcnnlex}{\nobreak{Lex. Neural Grid}}
\newcommand{\gridallnoun}{\nobreak{Grid-all nouns}}

\title{A Unified Neural Coherence Model}
\author{Han Cheol Moon$^*$$^\P$, Tasnim Mohiuddin$^{*\P}$, Shafiq Joty\thanks{*Equal contribution} $ ^\P$$^\S$, and Xu Chi$^\dag$\\
  $^\P$Nanyang Technological University, Singapore \\
  $^\S$Salesforce Research Asia, Singapore \\
  $^\dag$A*STAR, Singapore \\
$^\P$\texttt{\{hancheol001@e., mohi0004@e.\}ntu.edu.sg} \\ $^\S$\texttt{sjoty@salesforce.com}\\ $^\dag$\texttt{cxu@simtech.a-star.edu.sg}}

\date{}

\begin{document}
\maketitle
\begin{abstract}

Recently, neural approaches to coherence modeling have achieved state-of-the-art results in several evaluation tasks. However, we show that most of these models often fail on harder tasks with more realistic application scenarios. In particular, the existing models underperform on tasks that require the model to be sensitive to local contexts such as candidate ranking in conversational dialogue and in machine translation. In this paper, we propose a unified coherence model that incorporates sentence grammar, inter-sentence coherence relations, and global coherence patterns into a common neural framework. With extensive experiments on local and global discrimination tasks, we demonstrate that our proposed model outperforms existing models by a good margin, and establish a new state-of-the-art.


\end{abstract}

\section{Introduction}

Coherence modeling involves building text analysis models that can distinguish a coherent text from incoherent ones. It has been a key problem in discourse analysis with applications in text generation, summarization, and coherence scoring.

Various linguistic theories have been proposed to formulate coherence, some of which have inspired development of many of the existing coherence models. These include the   \textbf{entity-based} {local} models \cite{Barzilay:2008,Elsner:2011} that consider syntactic realization of entities in adjacent sentences,  inspired by the Centering Theory \cite{Grosz:1995}. Another line of research uses \textbf{discourse relations} between sentences to predict {local} coherence \cite{Pitler:2008,Lin:2011}. These methods are inspired by the discourse structure theories like Rhetorical Structure Theory (RST) \cite{Mann88} that formalizes coherence in terms of discourse relations. Other notable methods include \textbf{word co-occurrence} based {local} models \cite{Soricut:2006}, \textbf{content} (or topic distribution) based {global}  models \cite{barzilay-lee-2004}, and \textbf{syntax} based local and global models \cite{Louis:2012:CMB}.

With the neural invasion, some of the above traditional models have got their neural versions with much improved performance. For example, \citet{li-hovy:EMNLP20142} implicitly model syntax and inter-sentence relations using a neural framework that uses a recurrent (or recursive) layer to encode each sentence and a fully-connected layer with $\text{sigmoid}$ activations to estimate coherence probability for every window of three sentences. \citet{li-jurafsky:2017} incorporate global topic information with an encoder-decoder architecture, which is also capable of generating discourse. \newcite{mesgar-strube-2018-neural} model change patterns of salient semantic information between sentences.  \newcite{dat-joty:2017,joty-etal-2018-coherence} propose neural entity grid models using convolutions over distributed representations of entity transitions, and report state-of-the-art results in standard evaluation tasks on the Wall Street Journal (\textsc{wsj}) corpus.

Traditionally coherence models have been evaluated on two kinds of tasks. The first kind includes synthetic tasks such as \emph{discrimination} and \emph{insertion} that evaluate the models directly based on their ability to identify the right order of the sentences in a text with different levels of difficulty \cite{Barzilay:2008,Elsner:2011}. The latter kind of tasks evaluate the impact of coherence score as another feature in downstream applications like \emph{readability assessment} and \emph{essay scoring} \cite{Barzilay:2008,mesgar-strube-2018-neural}.

Although coherence modeling has come a long way in terms of novel models and innovative evaluation tasks \cite{Elsner:2011-chat,joty-etal-2018-coherence}, it is far from being solved. As we show later, state-of-the-art models often fail on harder tasks like \emph{local discrimination} and \emph{insertion} that ask the model to evaluate a local context (\eg\ a 3-sentence window). This task has direct applications in utterance ranking \cite{lowe-etal-2015-ubuntu} or bot detection\footnote{http://workshop.colips.org/wochat/data/index.html} in dialogue, and for sentence ordering in summarization.


According to \newcite{grosz-sidner-1986}, three factors \emph{collectively} contribute to discourse coherence: \Na the organization of discourse segments, \Nb intention or purpose
of the discourse, and \Nc attention or focused items. The entity-based approaches capture attentional structure, the syntax-based approaches consider intention, and the organizational structure is largely captured by models that consider discourse relations and content (topic) distribution. Although methods like \cite{elsner-etal-2007-unified,li-jurafsky:2017} attempt to combine these aspects of coherence, to our knowledge, no methods consider all the three aspects together in a single framework.

In this paper, we propose a unified neural model that incorporates sentence grammar (intentional structure),  discourse relations, attention and topic structures in a single framework. We use an LSTM sentence encoder with explicit language model loss to capture the syntax. Inter-sentence discourse relations are modeled with a bilinear layer, and a lightweight convolution-pooling is used to capture the attention and topic structures. We evaluate our models on both local and global discrimination tasks on the benchmark dataset. Our results show that our approach outperforms existing methods by a wide margin in both tasks. We have released our code at \href{https://ntunlpsg.github.io/project/coherence/n-coh-emnlp19/}{https://ntunlpsg.github.io/project/coherence/n-coh-emnlp19/} for research purposes. 

\section{Related Works}

Inspired by various linguistic theories of discourse, many coherence models have been proposed. In this section, we give a brief overview of the existing coherence models. 

Motivated by the Centering Theory \cite{Grosz:1995}, \citet{Barzilay:2005, Barzilay:2008} proposed the \textbf{entity-based} local model for representing and assessing text coherence, which showed significant improvements in two out of three evaluation tasks. Their model represents a text by a two-dimensional array called \textit{entity grid} that captures local transitions of discourse entities across sentences as the deciding patterns for assessing coherence. They consider the \textit{salience} of the entities to distinguish between transitions of important entities from unimportant ones, by measuring the occurrence frequency of the entities.

Subsequent studies extended the basic entity grid model. By including non-head nouns as entities in the grid, \citet{Elsner:2011} gained significant improvements. They incorporate entity-specific features like named entity, noun class, and modifiers to distinguish between entities of different types, which led to further improvements. Instead of using the transitions of grammatical roles, \citet{Lin:2011} model the transitions of \textbf{discourse roles} for entities.  \citet{Feng:2012} used the basic entity grid, but improved its learning to rank scheme. Their model learns not only from the original document and its permutations but also from ranking preferences among the permutations themselves.

\citet{Guinaudeau:2013} proposed a \textbf{graph-based} unsupervised method for modeling text coherence. Assuming the sentences in a coherent discourse should share the same structural syntactic patterns, \citet{Louis:2012:CMB} introduced a coherence model based on \textbf{syntactic patterns} in text. Their proposed method comprises of local and global coherence model, where the former one captures the co-occurrence of structural features in adjacent sentences and the latter one captures the global structure based on clusters of sentences with similar syntax. 

Our model also considers syntactic patterns through a biLSTM sentence encoder that is trained on an explicit language modeling loss. Compared to the entity grid and the syntax-based models, our model does not require any syntactic parser.  

With the \textit{neural tsunami}, some of the above traditional models have got their \textbf{neural} versions with better performance. \citet*{li-hovy:EMNLP20142} proposed a neural framework to compute the coherence score of a document by estimating coherence probability for every window of three sentences. \citet{li-jurafsky:2017} proposed two \textit{encoder-decoder} models, where the first model incorporates global discourse information (e.g., topics) by feeding the output of a sentence-level HMM-LDA model \cite{pmlr-v2-gruber07a} and the second model is trained end-to-end with variational inference. Our proposed model also models inter-sentence relations and global coherence patterns. We use a bi-linear layer to model relations between two consecutive sentences exclusively. Also, our global model implements a light-weight convolution that requires much less parameters, which gives better generalization. Moreover, we train the whole network end-to-end with a window-based adaptive pairwise ranking loss.   

\citet{dat-joty:2017} proposed a neural version of the entity grid model where they first transform the grammatical roles in a grid into their distributed representations. Then they employ a convolution operation over it to model entity transitions in the distributed space. Finally, they compute the coherence score from the convoluted features by a spatial max-pooling operation. The model is trained with a \textbf{document-level} (global) pairwise ranking loss. \citet{joty-etal-2018-coherence} improve the neural entity grid model by \textit{lexicalizing} its entity transitions They use \textit{off-the-shelf} word embeddings to achieve better generalization with the lexicalized model. As we will demonstrate, because of the spatial-pooling operation, entity-based neural models are not sensitive to  mismatch of \textbf{local patterns} in a document limiting their applicability to tasks that require local discrimination. Another crucial limitation of employing a \textbf{document-level} pairwise ranking loss is that the loss from the document-level permutation for a negative document may \textbf{penalize the convolution kernel weights} even if the local permutation matches that of the positive document. In contrast, we apply a {window-level} (local) \textbf{adaptive pairwise ranking loss} that gets activated only if the corresponding windows of the positive and negative documents differ. This way our model is sensitive to local patterns without penalizing the weights unfairly. We capture global patterns using a separate light-weight convolution module.



\section{Proposed Model}

\begin{figure*}[!ht]
	\begin{center}
		\includegraphics[scale=0.50]{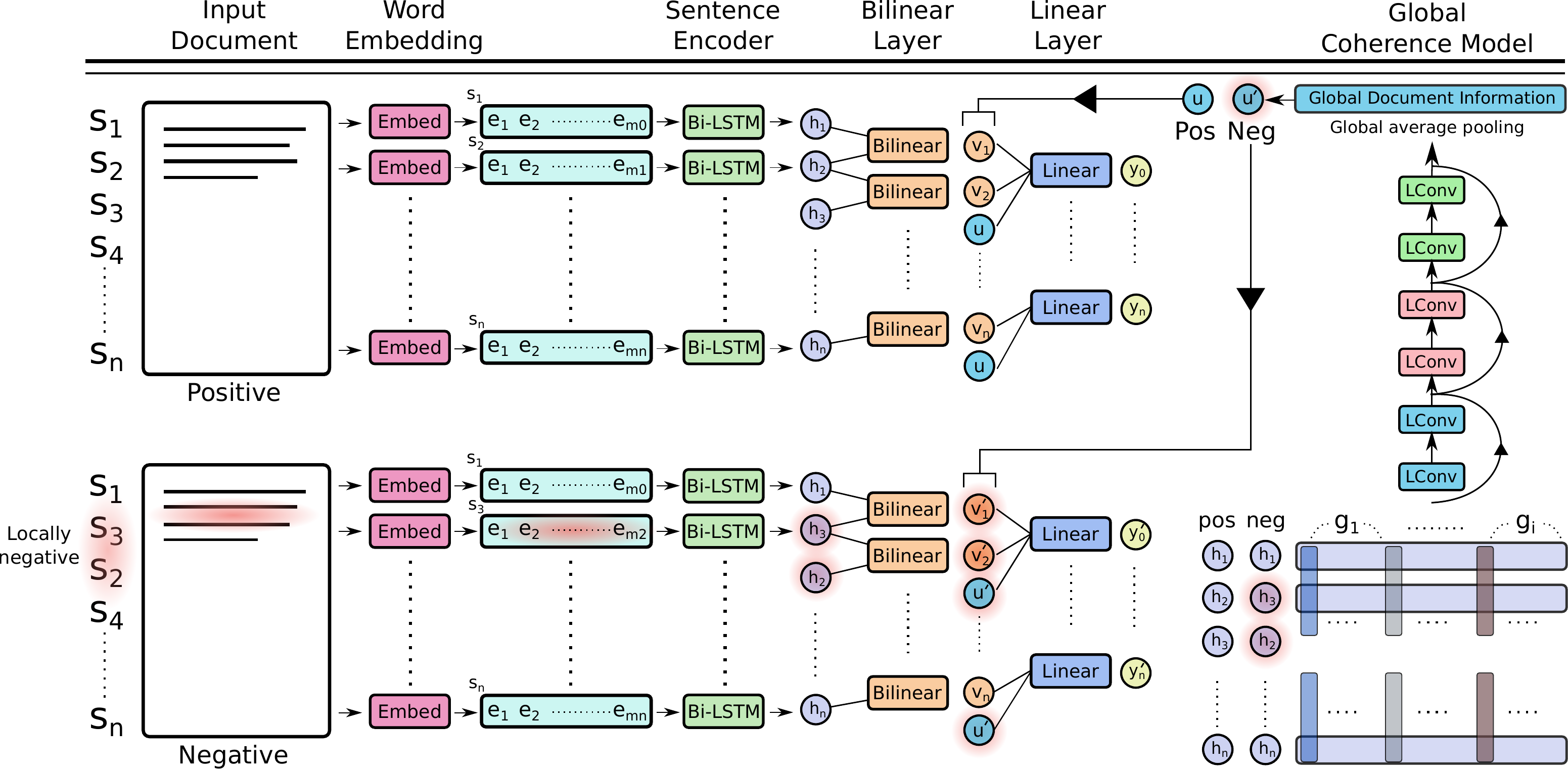}
	\end{center}
	\caption{An overview of the proposed coherence model (best viewed in color). The superscript `$'$' above output of each component denotes negative outputs and the red shade represents incoherent portions in the document. Note that all network parameters and components are shared regardless of the input documents.}
	\label{fig:bilinear}
\end{figure*}



Let ${D}= (\mathbf{s}_1, \cdots, \mathbf{s}_n )$ be a document consisting of $n$ sentences. Our goal is to assess its coherence score. Figure \ref{fig:bilinear} provides an overview of our proposed unified coherence model. It has four components in a Siamese architecture \cite{Bromley93}: \Ni a sentence encoder (\cref{subsec:enc}), \Nii a local coherence model (\cref{subsec:loc}), \Niii a global coherence model (\cref{subsec:glb}), and \Niv a coherence scoring layer (\cref{subsec:scr}). For encoding a sentence, we first map each word of the sentence to its corresponding vector representation. We then use a bidirectional LSTM sentence encoder with explicit \textbf{language model loss} to capture the sentence grammar. Given the sentence representations, the local and global coherence model extract the respective coherence features. The local coherence model implements a bilinear layer to model inter-sentence discourse relations. This layer captures the \textbf{local contexts} of the document. To capture the attention (entity distribution) and topic structures, i.e., the \textbf{global coherence} of the document, our global coherence model uses a \textit{light-weight convolution} \cite{payless_wu2018} with {average pooling}. The coherence scoring is a linear layer that evaluates the coherence from the extracted features. The whole architecture is trained \textit{end-to-end} with a pairwise ranking loss. In the following, we elaborate on different components of our proposed model.

\subsection{Modeling Intention} \label{subsec:enc}

A discourse has a purpose such as describing an event, explaining some results, evaluating a product, etc. As such, sentences in the discourse should support the purpose as a whole. The syntactic structure of the sentence can be used to model the intent structure \cite{Louis:2012:CMB}. We use a bidirectional long short-term memory or bi-LSTM  \cite{Hochreiter:1997} to encode each sentence into a vector representation while modeling its compositional structure. 

For an input sentence $\mathbf{s}_i = (w_1, \cdots, w_m)$ of length $m$, we first map each word $w_t$  to its corresponding vector representation $\mathbf{e}_t \in \mathbb{R}^{d}$, where $d$ is the dimension of the word embedding. The LSTM recurrent layer then computes a compositional representation ${\mathbf{h}}_t \in \mathbb{R}^{p}$ at every time step $t$ by performing nonlinear transformations of the current time step word vector representation $\mathbf{e}_t$ and the output of the previous time step ${\mathbf{h}}_{t-1}$, where $p$ is the number of features in the LSTM hidden state. The output of the last time step ${\mathbf{h}_m}$ is considered as the representation of the sentence. A bi-LSTM processes a given sentence $\mathbf{s}_i$ in two directions: from \textit{left-to-right} and  \textit{right-to-left}, yielding a representation $\mathbf{h}_i = [\overrightarrow{\mathbf{h}_m}; \overleftarrow{\mathbf{h}_m}] \in \mathbb{R}^{2p}$, where `;' denotes concatenation.

We train our sentence encoder with an explicit \textbf{language model loss}. A bidirectional language model combines a forward and a backward language model (LM). Similar to  \citet{Peters:2018}, we jointly minimize the \textit{negative log-likelihood} of the forward and backward directions:

\begin{eqnarray}
\mathcal{L}_{lm} = - \hspace{-0.1em} \sum_{j=1}^{m} \hspace{-0.5em} & \bigg( \hspace{-0.1em} \log p(w_{j} | w_{1}, \cdots, w_{j-1}; \theta,  \overrightarrow{\theta}_{\text{lstm}})  + \nonumber \\[-0.5em]  
 & \hspace{-0.8em}\log p(w_{j} | w_{j+1}, \cdots, w_{m}; \theta, \overleftarrow{\theta}_{\text{lstm}}) \bigg ) \label{eqn:lm-likelihood}
\end{eqnarray}
\normalsize

\noindent where $\overrightarrow{\theta}_{\text{{lstm}}}$ and $\overleftarrow{\theta}_{\text{{lstm}}}$ are the parameters of the forward and backward LSTMs, and $\theta$ denote the rest of the parameters which are shared.

\subsection{Modeling Inter-Sentence Relation} \label{subsec:loc}

Discourse relations between sentences reflect the organizational structure of a discourse that can be used to evaluate the coherence of a text \cite{Lin:2011,li-hovy:EMNLP20142}. To model inter-sentence discourse relations, we use a bilinear model. Our bi-LSTM sentence encoder gives a representation $\mathbf{h}_i \in \mathbb{R}^{2p}$ for each sentence $\mathbf{s}_i$ in the document.  We feed the representations of every two consecutive sentences $(\mathbf{h}_i, \mathbf{h}_{i+1})$ to this layer, which applies a bilinear transformation as: 

\begin{eqnarray}
\mathbf{v}_i = \mathbf{h}_i^T \mathbf{W}_{b} \mathbf{h}_{i+1} + \mathbf{b} 
\end{eqnarray}

\noindent where $\mathbf{W}_{b} \in \mathbb{R}^{q \times 2p \times 2p}$ is a learnable tensor, and $\mathbf{b}\in \mathbb{R}^{q}$ is a learnable bias vector. Here $q$ is the number of output features (i.e., $\mathbf{v}_i \in \mathbb{R}^q)$.


\subsection{Modeling Global Coherence Patterns} \label{subsec:glb}

The model proposed so far captures only local information. However, global discourse phenomena like entity or topic distributions are also important for coherence evaluation \cite{barzilay-lee-2004,elsner-etal-2007-unified,Louis:2012:CMB}. Global coherence is modeled in our architecture by a \textit{convolution-pooling} mechanism.



As shown in the Figure \ref{fig:bilinear}, our \textbf{global coherence} sub-module takes the representations $\mathbf{H} = (\mathbf{h}_1, \cdots , \mathbf{h}_n)$ of all the sentences in a document  generated by the bi-LSTM encoder. The module uses six convolution layers with residual connections, followed by an average pooling layer. Instead of using regular convolutions, we use \textit{light-weight convolution} \cite{payless_wu2018}, which is built upon \emph{depth-wise} convolution \cite{chollet-2016}.

\begin{figure}[t!]
\includegraphics[width=\linewidth]{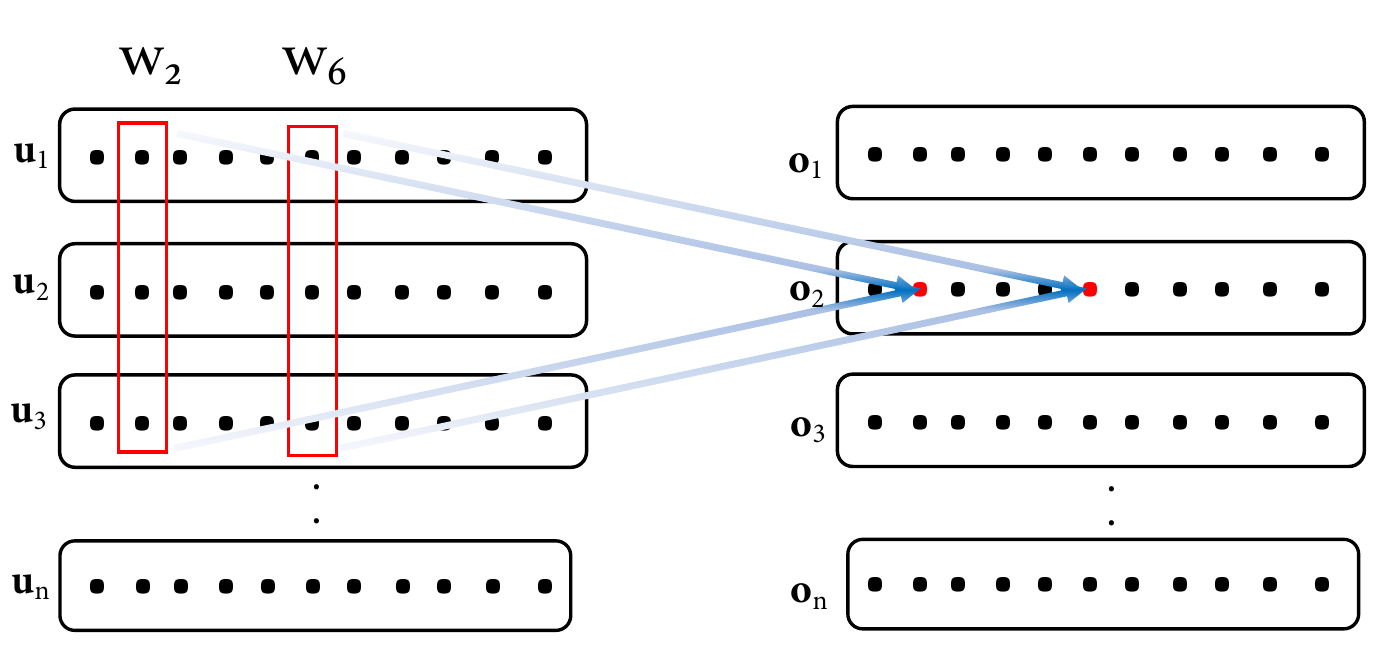}
\caption{\small Depth-wise convolution for kernel size $k=3$. The convolutions are done over the input dimensions.}
\label{fig:dc}
\end{figure}

\noindent \textbf{Depth-wise convolutions} perform a convolution independently over every input channel which significantly reduces the number of parameters as shown in Figure \ref{fig:dc}. For a given input $\mathbf{H} \in \mathbb{R}^{n \times d}$, the output $\mathbf{O} \in \mathbb{R}^{n \times d}$  of the depth-wise convolution  with convolution weight $\mathbf{W} \in \mathbb{R}^{d \times k}$ with kernel size $k$ for element $i$ and output dimension $c$ can be written as:

\begin{align}
\mathbf{O}_{i,c} &= \text{Dcon}(\mathbf{H}, \mathbf{W}_{c,:},i,c) \nonumber \\ &= \sum_{j=1}^{k} \mathbf{W}_{c,j} \mathbf{H}_{(i+j-\lceil \frac{k+1}{2}\rceil),c} 
\end{align}

\noindent Compared to the regular convolutions,depth-wise convolutions reduces the number of parameters  from $d^2k$ to $dk$ (note that $d=2p$ in our case).

\noindent \textbf{Light-weight Convolutions} make the depth-wise convolution even simpler by sharing groups of output channels and normalizing weights across the temporal dimension using a softmax. It has a fixed context window which determines the importance of context elements with a set of weights that do not change over time steps. For the $i$-th element in the sequence and output channel $c$, \textit{light-weight convolution} computes:

\begin{align}
\text{Lcon}(\mathbf{H}, &\mathbf{W}_{g,:}, i, c) \nonumber \\ 
&= \text{Dcon}(\mathbf{H}, \text{softmax}(\mathbf{W}_{g,:}), i, c)
\end{align}

\noindent where $g = \lceil \frac{cG}{d}\rceil$ with $G$ being the number of groups. The number of parameters with light-weight convolutions reduces to $H.k$. \citet{payless_wu2018} show that models equipped with {light-weight convolution} exhibit better generalization compared to regular convolutions. It is indeed crucial in our case since we use convolutions to model a document, with large kernel size it would be difficult to learn from small datasets compared to (sentence-level) machine translation datasets.


The light-weight convolution layers generate $d$ feature maps $\mathbf{f}^i\in\mathbb{R}^{n}, i=1,...,d$ for each input document by the convolutional operation over an input dimension across all the sentences in a document. Subsequently, \textbf{global average pooling} is performed over the extracted feature maps to achieve a global view of the input document. The achieved global feature $\mathbf{u} \in \mathbb{R}^{d}$ can be expressed as follows:  

\begin{align}
\mathbf{u} = \frac{1}{n}[\mathbf{1}\cdot \mathbf{f}^1, ... , \mathbf{1}\cdot \mathbf{f}^d]
\end{align}

\noindent where $\mathbf{1}\in \mathbb{R}^{n}$ is the vector of ones and $n$ is the number of sentences in an input document. The global document level features  $\mathbf{u}$ are then concatenated with the local features of each consecutive sentence pair ($\mathbf{v}_i$; $\mathbf{v}_{i+1})$ in the document, i.e., the output of the bilinear layer (see Figure \ref{fig:bilinear}). 








\begin{align}
\mathbf{z}_i = [\mathbf{v}_i; \mathbf{v}_{i+1}; \mathbf{u}]
\end{align}
where $\mathbf{z}_i \in \mathbb{R}^{2q+2p}$ and `;' denotes concatenation.

\subsection{Coherence Scoring} \label{subsec:scr}

We then feed the concatenated global and local features $\mathbf{z}_i$ to the final linear layer of our model to compute the coherence score ${y}_i \in \mathbb{R}^{n}$ for each local window.

\begin{align}
{y}_i = \mathbf{z}_i^T \mathbf{w}_l + {b}_{\text{l}}
\end{align}

\noindent where $\mathbf{w}_l $ is weight vector and $b_l$ is a bias. {The final decision on documents is made by summing up all local scores of documents and compares the summed scores.}

\subsection{Overall Objective and Training Details}

Our model assigns a coherence score $y_i$ to every possible local window $D^{\ell}$ in the document $D$, where $\ell$ is the \textit{local window} index. During implementation, the input document is \textit{padded}, so that the number of possible local window is the same as the number of sentences ($n$) in the document $D$. 


Let $\mathbf{y} = \Omega({{D}}|\Theta)$ define our model that produces the coherence scores $\mathbf{y} = (y_1, \ldots, y_n)$ for an input document ${D}$, with $\Theta$ being the parameters. We use a \textit{window-level pairwise ranking approach} \cite{collobert2011natural} to learn $\Theta$. 



Our training set contains \textit{ordered} pairs of documents $({D_{\text{pos}}}, {D_{\text{neg}}})$, where document ${D_{\text{pos}}}$ exhibits a higher degree of coherence than document ${D_{\text{neg}}}$. See Section \ref{sec:eval} for details about the dataset. 
We seek to learn $\Theta$ that assigns higher coherence scores to ${D_{\text{pos}}}$ than to ${D_{\text{neg}}}$. {We observed that the naive pairwise ranking loss that uses a fixed margin unfairly penalizes the locally positive sentences during training. In other words, the loss should be active only for local windows that differ in ${D_{\text{pos}}}$ and ${D_{\text{neg}}}$.} To address this, we propose to use an \textbf{adaptive pairwise ranking loss} {$\mathcal{L}_\Theta$} defined as follows.

\begin{eqnarray}
\mathcal{L}_\Theta = \frac{1}{n} \sum_{\ell=1}^n & \max \left\{ 0, \phi ({D_{\text{pos}}^{\ell}}, {D_{\text{neg}}^{\ell}}) - \right. \nonumber \\ 
& \left. \Omega({D}_{\text{pos}}^{\ell}|\Theta) + \Omega({D}_{\text{neg}}^{\ell}|\Theta) \right\}
\end{eqnarray}

\noindent {where $\phi ({D_{\text{pos}}^{\ell}}, {D_{\text{neg}}^{\ell}})$ is  an adaptive margin given by}

\begin{align*}
\phi ({D_{\text{pos}}^{\ell}}, {D_{\text{neg}}^{\ell}}) = 
\begin{cases}
0 & \textrm{if } {D_{\text{pos}}^{\ell}} = {D_{\text{neg}}^{\ell}}\\
\tau & \textrm{otherwise}
\end{cases}
\end{align*}
where $\tau$ is a margin constant. 

\noindent Our total loss, $\mathcal{L}_\Theta = \mathcal{L}_\Theta + \mathcal{L}_{lm}$. 

Note that our model shares all the layers and components, i.e., $\Theta$ to obtain $\Omega({D_{\text{pos}}}|\Theta)$ and $\Omega({D_{\text{neg}}}|\Theta)$ from a pair of document $({D_{\text{pos}}}, {D_{\text{neg}}})$. Therefore, once trained, it can be used to score any input document independently.





\section{Evaluation Tasks and Datasets} \label{sec:eval}

\begin{figure}[t]
	\begin{center}
		\includegraphics[scale=0.35]{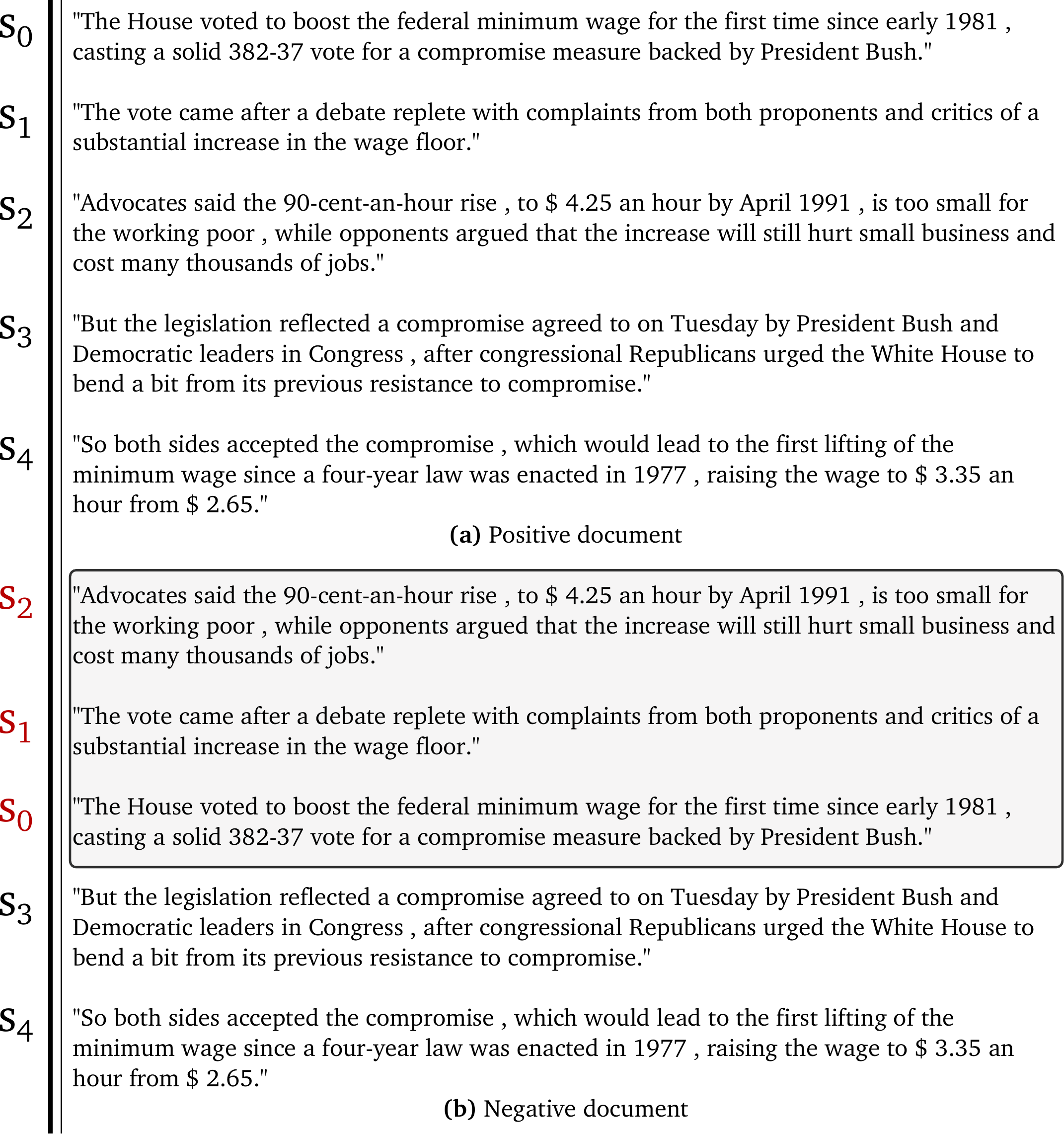}
	\end{center}
	\caption{Sample data in the local permutation data set. (a) is the positive document, WSJ0098, and (b) is the negative version of the positive document, which is locally sentence-order permuted.}
	\label{fig:sample_data}
\end{figure}

For comparison purposes with previous work, we evaluate our models on the standard \textbf{``global'' discrimination} task \cite{Barzilay:2008},  where a document is compared to a \textit{random permutation} of its sentences, which is considered to be incoherent. We also evaluate on an \emph{inverse discrimination} task  \cite{joty-etal-2018-coherence}, where the sentences of the original document are placed in the reverse order to create the incoherent document. Similar to them, we do not train our models explicitly on this task, rather we use the trained model from the standard discrimination task.  In addition and more importantly, we evaluate the models on a more challenging \textbf{``local'' discrimination} task, where two documents differ only in a local context (\eg\ a 3-sentence window), as shown with an example in Figure \ref{fig:sample_data}.

\paragraph{Dataset for Global Discrimination.} 

We follow the same experimental setting of the \textsc{WSJ} news dataset as used in previous works  \cite{joty-etal-2018-coherence,dat-joty:2017,Elsner:2011,Feng:2014}. Similar to them, we use 20 random permutations of each document for both training and testing, and exclude permutations that match the original one. 
Table \ref{table:global_dataset} summarizes the data sets used in global discrimination task.  We randomly selected 10\% of the training set for development purposes.

\begin{table}[tb!]
\centering
	\resizebox{0.8\columnwidth}{!}{%
		\begin{tabular}{l|l|c|c}
			\toprule
			& Sections & \# Doc. & \# Pairs   \\
			\midrule
			{Train} & {00-13}  & 1,378 & 26,422 \\
			{Test}  & {14-24}  & 1,053 & 20,411 \\
			\bottomrule
		\end{tabular}
	}
	\caption{Statistics of the \textsc{WSJ} news dataset used for \textbf{``global'' discrimination} task.}
	\label{table:global_dataset}
\end{table}



\begin{table}[t!]
	\resizebox{1.0\columnwidth}{!}{%
		\begin{tabular}{l|l|c|cccc}
			\toprule
			& Sections & \# Doc. & \multicolumn{4}{c}{\# Pairs}   \\
			&		   &		& $\mathcal{D}_{w=1}$ & $\mathcal{D}_{w=2}$ & $\mathcal{D}_{w=3}$ & $\mathcal{D}_{w=1,2,3}$\\
			\midrule
			{Train} & {00-13}  & 748 & 7,890 & 12,280 & 12,440 & 32,610\\
			{Test}  & {14-24}  & 618 & 6,568 & 9,936 & 9,906 & 26,410 \\
			\bottomrule
		\end{tabular}
	}
	\caption{Statistics on the \textsc{WSJ} news dataset used for \textbf{``local'' discrimination} task. The $w$ denotes the number of permuted local windows in a document.}
	\label{table:local_dataset}
\end{table}

\paragraph{Dataset for Local Discrimination.}
We use the same  WSJ articles used in the global discrimination task (Table \ref{table:global_dataset}) to create our local discrimination datasets. Sentences inside a local \textbf{window of size 3} are re-ordered to form a locally incoherent text. Only articles with \textbf{more than 10 sentences} are included in our dataset. This gives 748 documents for training and 618 for testing.  

We first set $w$ as the number of local windows that we want to permute in a document. Based on this, we create four datasets for our local discrimination task: $\mathcal{D}_{w=1}$, $\mathcal{D}_{w=2}$, $\mathcal{D}_{w=3}$ and $\mathcal{D}_{w=1,2,3}$. $\mathcal{D}_{w=1}$ contains the documents, where only one randomly selected window is permuted. Similarly, $\mathcal{D}_{w=2}$ contains the documents, where two randomly selected windows are permuted. $\mathcal{D}_{w=3}$ is similarly created for 3 windows. $\mathcal{D}_{w=1,2,3}$ denotes the concatenated datasets. The number of negative documents for each article was restricted \textbf{not to exceed 20 samples}. Additionally, we exclude the cases of the overlap between windows. In other words, the sentences are allowed to be permuted only inside their respective window.

We randomly select 10\% of the training set for development purposes. Table \ref{table:local_dataset} summarizes the datasets. Consequently, the training and the test dataset for $\mathcal{D}_{w=1,2,3}$ consists of 32,610 and 26,410 pairs, respectively. 




\section{Experiments}


This section presents details of our experiment procedures and results. 

\subsection{Models Compared}

{We compare our proposed unified coherence model with several existing models. Some of the baselines that are not publicly available were re-implemented during experiments, otherwise we conducted experiments with publicly available codes, and the rest of the reported results are from their original papers. In the following sections, we present brief descriptions of the existing models.}

\paragraph{Distributed Sentence Model (L\&H).} This is the neural model proposed by  \citet{li-hovy:EMNLP20142}. Similar to our local model, it extracts local coherence features for small windows of sentences to compute the coherence score of a document. First, they use a recurrent or a recursive neural network to compute the representation for each sentence in the local window from its words and their pretrained embeddings. Then the concatenated vector is passed through a non-linear hidden layer, and finally the output layer decides if the window of sentences is a coherent text or not. The main differences between our implementation and the implementation referred in their paper are that we used a bi-LSTM (as opposed to simple RNN) for sentence encoding and trained the network with the Adam optimizer (as opposed to AdaGrad).

\paragraph{Grid-all nouns \& Extended grid (E\&C)\footnote{\href{https://bitbucket.org/melsner/browncoherence/src}{https://bitbucket.org/melsner/browncoherence/src}}.} \citet{Elsner:2011} report significant gains by including all nouns as entities in the original entity grid model as opposed to considering only head nouns. In their extended grid model, they used 9 additional  entity-specific features, 4 of which are computed from external corpora. 

\paragraph{Neural Grid \& Ext. Neural Grid (N\&J)\footnote{\href{https://github.com/datienguyen/cnn\_coherence}{https://github.com/datienguyen/cnn\_coherence}}.} These are the neural versions of the entity grid models as proposed by \cite{Joty17}. They use convolutions over grammatical roles to model entity transitions in the distributed space. In the extended model, they incorporate three entity-specific features.

\paragraph{Lex. Neural Grid (M\&J)\footnote{\href{https://github.com/taasnim/conv-coherence}{https://github.com/taasnim/conv-coherence}}.} \citet{joty-etal-2018-coherence} improved the neural grid model by lexicalizing the entity transitions. Experiment results for this model were obtained with the optimal setting described in the original paper.




 


\paragraph{Global Coherence Model.}
This is the global coherence model component in our proposed unified model as described in Section \ref{subsec:glb}. The model extracts document-level features through lightweight convolutions. The extracted features are subsequently averaged along the temporal dimension, which is in turn used in a linear layer for coherence scoring. This model used a kernel size of 5 and each document was padded by the size of 3. 

\subsection{Settings of Our Model}

We held out 10\% of the training documents to form a development set (DEV) on which we tune the hyper-parameters of our models. In our experiments, we use both \texttt{word2vec} \cite{Mikolov.Sutskever:13} and \texttt{ELMo} \cite{Peters:2018} for the distributed representations of the words. Unlike \texttt{word2vec}, \texttt{ELMo} is capable of capturing both subword information and contextual clues. We implemented our models in PyTorch framework on a Linux machine with a single GTX 1080 Ti GPU.


During training, for optimization we use Adam optimizer \cite{KingmaB14} with $L_2$ regularization (0.00001 regularization parameter). We trained the model up to 25 epochs to make the models' performance converge. 
To search for optimal parameters, we conducted various experiments while varying the hyper-parameters. Precisely, minibatch size in \{5, 10, 20, 25\}, sentence embedding size in \{128, 256\}, lightweight convolution kernel size in \{3, 5, 7, 9\}, bilinear output dimension size in \{32, 64\} are investigated. We present the optimal hyper-parameter values in the \textit{supplementary document}. The results are reported  by averaging over five different runs of the model with different seeds for statistical stability. 

\subsection{Results on Local Discrimination}

Table \ref{table:local-ordering} shows the results in accuracy on the \textbf{``local'' discrimination} task. From the table, we see that existing models including our global model perform poorly compared to our proposed local models. They are likely to fail in distinguishing the text segments that are locally coherent and penalize them unfairly. One of the possible explanations of this phenomenon can be found in the nature of the global model. These models (except L\&H) are designed to make a decision at a global level, thus they are likely to penalize locally coherent segments of a text. This observation is further bolstered by the performance of our local coherence models, which show higher sensitivity in discriminating locally coherent texts and achieve significantly higher accuracy compared to the baseline models and our global model.

\begin{table}[t!]
	\centering
	\scalebox{0.62}{
		\begin{tabular}{l|lcccc}
			\toprule 
			\textbf{Model} & \textbf{Emb.} &  $\mathcal{D}_{w=1,2,3}$ &  $\mathcal{D}_{w=1}$ &  $\mathcal{D}_{w=2}$ &  $\mathcal{D}_{w=3}$ \\
			\midrule
			\gridcnnlex\ (M\&J)$^\star$  & \texttt{word2vec} & 60.27 & 56.11 & 60.23 & 62.23  \\
			\gridcnnlex\ (M\&J) & \texttt{word2vec} & 55.01 & 53.81 & 55.37 & 56.16\\
			Dist. sentence (L\&H) & \texttt{word2vec} & 6.76 & 4.28 & 6.82 &9.25\\
			\midrule
			{Our Global Model}  & \texttt{word2vec} & 57.24 & 53.35 & 56.58 & 59.67 \\
			\midrule
			{Our Local Model} & \texttt{word2vec} & {73.23} & {{66.21}} & {73.16} & {77.93} \\
			{Our Local Model} & \texttt{ELMo}   & 74.12 & 65.82 & {73.54} & 78.16  \\
			\midrule
			{Our Full Model}     & \texttt{word2vec} & 75.37 & \textbf{67.29} & 75.58 & 80.21 \\
			{Our Full Model}     & \texttt{ELMo} & \textbf{77.07} & 64.38 & \textbf{76.12} & \textbf{81.23}  \\
			\bottomrule 
		\end{tabular}
	}
	\caption{Results in accuracy on the \textbf{Local {Dis}crimination task}. $^\star$ is a pre-trained model on the global discrimination task.} 
	\label{table:local-ordering} 
\end{table}

Another aspect to notice here is that the performance of all the models become gradually better with the increase in the number of permutation windows in the dataset. This is not surprising because in the datasets with a lower number of permutation windows, the difference between a positive and a negative document is very subtle. For example, in $\mathcal{D}_{w=1}$ dataset, positive and negative documents differ only in a small window position. Another interesting observation regarding the entity-grid based neural models is that the model pretrained on the global discrimination task performs better than the ones trained on the specific tasks. From the table, we observe that our full model with \texttt{ELMo} word embeddings achieves the highest accuracies on the datasets $\mathcal{D}_{w=1,2,3}$, $\mathcal{D}_{w=2}$, and $\mathcal{D}_{w=3}$ , while on the $\mathcal{D}_{w=1}$ dataset, our full model with the pretrained \texttt{word2vec} embeddings performs the best. {The reason could be that with more generalized contextual embeddings, our model losses the discrimination capability for small changes in the document.}

\subsection{Results on Global Discrimination}

Table \ref{table:ordering} presents the results in accuracy on the two \textbf{``global'' discrimination} tasks -- the \textit{Standard} and the \textit{Inverse} order discrimination. {The reported results of the entity-grid models are from the original papers. `\gridcnnlex\ (M\&J)(code)' refers to the results achieved by running the code released by \citet{joty-etal-2018-coherence} on our machine.}

From the table, we see that our unified neural coherence model outperforms the existing models by a good margin. In this dataset, our best model with the \texttt{word2vec} embeddings achieves 90.42\% and 95.27\%, on \textit{Standard} and \textit{Inverse} order discrimination tasks, respectively. We achieve the best results with our proposed model by using the \texttt{ELMo} word embeddings, where we get 93.19\% and 96.78\%  accuracies on \textit{Standard} and  \textit{Inverse} order discrimination tasks, respectively. 

\begin{table}[tb!]
	\resizebox{1.0\columnwidth}{!}{%
		\begin{tabular}{cl|ccc}
			\toprule 
			& \textbf{Model} & \textbf{Emb.} & \textbf{Standard}  & {\textbf{Inverse}} \\
			\midrule 
			\multirow{1}{*}{I} 
			& Dist. sentence (L\&H) & \texttt{word2vec} & 17.39 & 18.11 \\
			\midrule
			\multirow{2}{*}{II} 
			& \gridallnoun\ (E\&C) & - & 81.60 & 75.78  \\
			& Ext. Grid (E\&C) & - & 84.95 & 80.34  \\
			\midrule
			\multirow{2}{*}{III} 
			& \gridcnn\ (N\&J) & \texttt{Random} &  84.36 & 83.94 \\
			& Ext. \gridcnn\ (N\&J) & \texttt{Random} & {85.93} & 83.00 \\
			\midrule 
			\multirow{3}{*}{IV} 
			& \gridcnnlex\ (M\&J) & \texttt{Random}  & 87.03 & 86.88 \\ 
			& \gridcnnlex\ (M\&J)(paper) & \texttt{word2vec}  &  {88.56}& {88.23}\\
			& \gridcnnlex\ (M\&J)(code) & \texttt{word2vec}  &  88.51 & 88.13\\
			\midrule 
			\multirow{2}{*}{V} 
			& {Our Best Model} & \texttt{word2vec} & 90.42 & 95.27 \\
			& {Our Best Model} & \texttt{ELMo} & \textbf{93.19} & \textbf{96.78} \\

			\bottomrule 
		\end{tabular}
	}
	\caption{Results in accuracy on the \textbf{Global {Dis}crimination} task.}
	\label{table:ordering}
\end{table}

\subsection{Ablation Study}

To investigate the impact of different components in our proposed model, we conducted two sets of ablation study on the local and global discrimination tasks. Specifically, we want to see: \Ni the impact of our global model component, and \Nii the impact of the language model (LM) loss.

\paragraph{Local Discrimination.}
In the local discrimination task,  we first compare the performance of the proposed model without the LM loss. As shown in the first block of Table \ref{table:ablation}, addition of the global model to the local model degrades the performance on the $\mathcal{D}_{w=1}$ dataset  by 1.17\% and 1.21\% for \texttt{word2vec} and \texttt{ELMo} embeddings, respectively. While for the other datasets, we see improvements in performance for the addition of the global model. However, in the presence of the LM loss (second block in Table \ref{table:ablation}), the addition of the global model improves the performance across all the datasets on the local discrimination task.

On the other hand, the addition of the  LM loss to our model (with/without global model) increases the accuracy in most of the datasets and embeddings. Exception is the \texttt{ELMo} embeddings on $\mathcal{D}_{w=1}$ dataset, where the overall performance drops by 1.60\% and 0.23\% for the local model with and without the global model, respectively.

Another interesting observation on $\mathcal{D}_{w=1}$ dataset is that in all the cases  \texttt{word2vec} embeddings outperforms \texttt{ELMo}. This unusual behavior of $\mathcal{D}_{w=1}$ dataset to the rests is not surprising because it is the hardest dataset where the difference between the positive and the negative document is subtle.  In this case, generally flexible and simple models outperform complex ones.

For the performance degradation of the global model in $\mathcal{D}_{w=1}$ case, we assume that in some texts, the global model fails to capture the significant feature from the locally negative region. Subsequently, the global feature is added into the score calculation at every local window, so the overall influence of the global model becomes bigger than that of the local model in the decision making.

\begin{table}[t!]
	\centering
	\scalebox{0.65}{
		\begin{tabular}{l|lcccc}
			\toprule 
			\textbf{Model} & \textbf{Emb.} &$\mathcal{D}_{w=1,2,3}$ &  $\mathcal{D}_{w=1}$ &  $\mathcal{D}_{w=2}$ &  $\mathcal{D}_{w=3}$ \\
			\midrule
			\textbf{Without LM Loss} \\
			{Local Model}  & \texttt{word2vec} & {73.23} & {{66.21}} & {73.16} & {77.93} \\
			{Local Model}  & \texttt{ELMo} & 74.12 & 65.82 & {73.54} & 78.16\\
			{Local + Global} Model  & \texttt{word2vec} & 74.69  & 65.04  & 75.27  & {79.69} \\
			{Local + Global} Model  & \texttt{ELMo} & {76.01} & 64.61 & 75.22 & 79.37\\
			\midrule
			\textbf{With LM Loss} \\
			{Local Model} & \texttt{word2vec} & 75.03 & {66.95} & 75.04 & {80.07} \\
			{Local Model} & \texttt{ELMo}   & {75.20} & 64.22 & {75.93} & 80.57  \\
			{Local + Global} Model     & \texttt{word2vec} & 75.37 & \textbf{67.29} & 75.58 & 80.21 \\
			{Local + Global} Model     & \texttt{ELMo} & \textbf{77.07} & 64.38 & \textbf{76.12} & \textbf{81.23}  \\
			\bottomrule 
		\end{tabular}
	}
	\caption{Ablation study of different model components on the \textbf{Local {Dis}crimination task}. 
	} 
	\label{table:ablation} 
\end{table}

\paragraph{Global Discrimination.}
We also studied the impact of our global model and the LM loss in the global discrimination task. As shown in Table \ref{table:ablation_global}, the addition of the global model and LM loss to the local model improves performance on the standard discrimination task by 1.34\%. 

However, the addition of global model impacts negatively on the inverse order task and degrades accuracies by 2.42\% and 1.28\% in the presence and absence of LM loss respectively. We suspect that the global model is adding noise because of the pooling operation, which throws away the spatial relation between sentences and provides the global information that is invariant to the sentence-order. But in this task, order information is crucial. In the inverse order task, we get the best performance by adding the LM loss to our local model.

\begin{table}[t!]
	\centering
	\scalebox{0.65}{
		\begin{tabular}{l|ccc}
			\toprule 
		 \textbf{Model} & \textbf{Emb.} & \textbf{Standard}  & {\textbf{Inverse}} \\
			\midrule
		 {Our Local Model} & \texttt{word2vec} & 88.93 & 94.72 \\
			 \quad + LM Loss & \texttt{word2vec} & 89.92 & \textbf{96.24}\\
			 \quad + Global Model & \texttt{word2vec} & 89.53  & 93.44\\
			 \quad + Global Model + LM Loss & \texttt{word2vec} & \textbf{90.27} & {93.82}\\
			 \qquad  (\textbf{Our Full Model}) & & &\\
			\bottomrule 
		\end{tabular}
	}
	\caption{Ablation study of different model components on the \textbf{Global {Dis}crimination task}.} 
	\label{table:ablation_global} 
\end{table}

\section{Conclusion}

In this paper, we proposed a unified coherence model. The proposed model incorporates a local coherence model and a global coherence model to capture the sentence grammar (intentional structure),  discourse relations, attention and topic structures in a single framework. The unified coherence model shows state of the art results on the standard coherence assessment tasks: the inverse-order and the global discrimination tasks. Also, our evaluation of the local discrimination task demonstrates the effectiveness of the unified coherence model in assessing global and local coherence of texts. 

\section*{Acknowledgments}

We would like to thank the anonymous reviewers for their comments. Shafiq Joty would like to thank the funding support from his Start-up Grant (M4082038.020). Also, this work is partly supported by SIMTech-NTU Joint Laboratory on Complex Systems.

\bibliography{coherence}
\bibliographystyle{acl_natbib}

\end{document}